\documentclass[10pt,twocolumn,letterpaper]{article}

\usepackage{iccv}
\usepackage{times}
\usepackage{epsfig}
\usepackage{graphicx}
\usepackage{amsmath}
\usepackage{amssymb}

\usepackage[pagebackref=true,breaklinks=true,letterpaper=true,colorlinks,bookmarks=false]{hyperref}

 \iccvfinalcopy 

\def\iccvPaperID{1583} 

\ificcvfinal\fi
\begin{document}

\title{Hide-and-Seek: Forcing a Network to be Meticulous for\\Weakly-supervised Object and Action Localization}

\author{Krishna Kumar Singh and Yong Jae Lee\\
	University of California, Davis}

\maketitle

\begin{abstract}
\vspace*{-0.1in}
We propose `Hide-and-Seek', a weakly-supervised framework that aims to improve object localization in images and action localization in videos.  Most existing weakly-supervised methods localize only the most discriminative parts of an object rather than all relevant parts, which leads to suboptimal performance. Our key idea is to hide patches in a training image randomly, forcing the network to seek other relevant parts when the most discriminative part is hidden. Our approach only needs to modify the input image and can work with any network designed for object localization.  During testing, we do not need to hide any patches.  Our Hide-and-Seek approach obtains superior performance compared to previous methods for weakly-supervised object localization on the ILSVRC dataset.  We also demonstrate that our framework can be easily extended to weakly-supervised action localization.
\end{abstract}

\vspace*{-0.1in}
\section{Introduction}
\vspace*{-0.05in}

Weakly-supervised approaches have been proposed for various visual classification and localization tasks including object detection~\cite{weber-eccv2000,fergus-cvpr2003,Crandall-ECCV2006,siva-eccv2012,bilen-bmcv2014,wang-eccv2014,song-nips2014,cinbis-arxiv2015,Oquab-cvpr15,zhou-cvpr2016,krishna-cvpr2016}, semantic segmentation~\cite{pathak-ICCV2015,khoreva-cvpr2016} and visual attribute localization~\cite{berg-eccv2010,wang-cvpr2013b,Xiao-iccv2015,Wang-CVPR2016,krishna-eccv2016}.  The main advantage of weakly-supervised learning is that it requires less detailed annotations compared to the fully-supervised setting, and therefore has the potential to use the vast weakly-annotated visual data available on the Web.  For example, weakly-supervised object detectors can be trained using only image-level labels (`dog' or `no dog') without any object location annotations.

Existing weakly-supervised methods identify discriminative patterns in the training data that frequently appear in one class and rarely in the remaining classes.  This is done either explicitly by mining discriminative image regions or features~\cite{weber-eccv2000,fergus-cvpr2003,Crandall-ECCV2006,siva-eccv2012,bilen-bmcv2014,song-icml2014,song-nips2014,cinbis-arxiv2015,krishna-cvpr2016} or implicitly by analyzing the higher-layer activation maps produced by a deep network trained for image classification~\cite{simonyan-iclr2014,Oquab-cvpr15,zhou-cvpr2016}.  However, due to intra-category variations or relying only on a classification objective, these methods often fail to identify the entire extent of the object and instead localize only the most discriminative part.

\begin{figure}[t!]
\centering
    \includegraphics[width=0.45\textwidth]{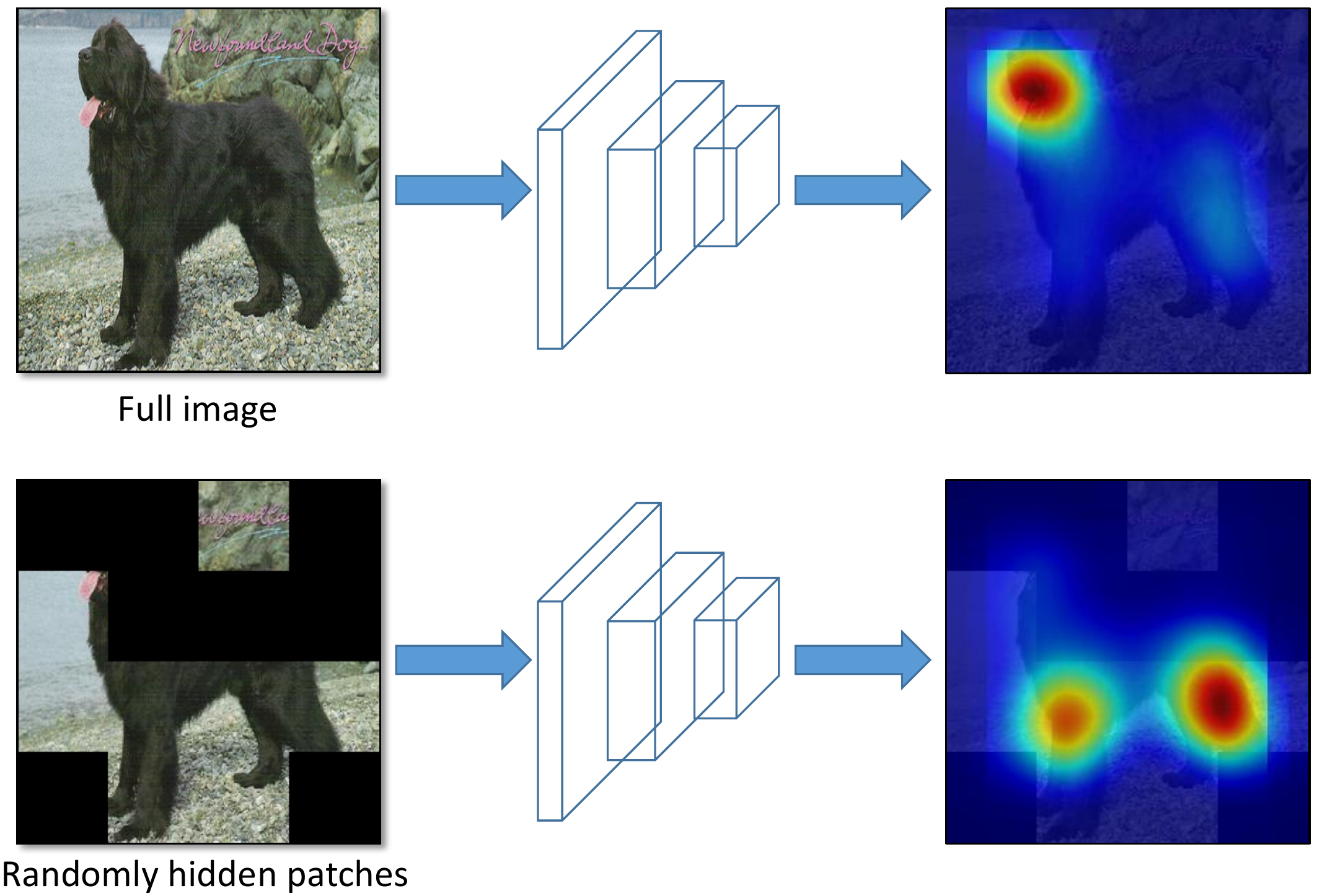}
    \caption{\textbf{Main idea.}  (Top row) A network tends to focus on the most discriminative parts of an image (e.g., face of the dog) for classification. (Bottom row) By hiding images patches randomly, we can force the network to focus on other relevant object parts in order to correctly classify the image as 'dog'.}
    \label{fig:teaser}
    \vspace*{-0.1in}
\end{figure}

Recent work tries to address this issue of identifying only the most discriminative part.  Song et al.~\cite{song-nips2014} combine multiple co-occurring discriminative regions to cover a larger extent of the object.  While multiple selections ensure larger coverage, it does not guarantee selection of less discriminative patches of the object in the presence of many highly discriminative ones.  Singh et al.~\cite{krishna-cvpr2016} use motion cues and transfer tracked object boxes from weakly-labeled videos to the images.  However, this approach requires additional weakly-labeled videos, which may not always be available.  Finally, Zhou et al.~\cite{zhou-cvpr2016} replace max pooling with global average pooling after the final convolution layer of an image classification network.  Since average pooling aggregates activations across an entire feature map, it encourages the network to look beyond the most discriminative part (which would suffice for max pooling).  However, the network can still avoid finding less discriminative parts if identifying a few highly-discriminative parts can lead to accurate classification performance, as shown in Figure~\ref{fig:teaser}(top row).

\vspace{-10pt}
\paragraph{Main Idea.} In this paper, we take a radically different approach to this problem.  Instead of making algorithmic changes~\cite{song-nips2014,zhou-cvpr2016} or relying on external data~\cite{krishna-cvpr2016}, we make changes to the \emph{input image}.  The key idea is to \emph{hide} patches from an image during training so that the model needs to \emph{seek} the relevant object parts from what remains.  We thus name our approach `Hide-and-Seek'.  Figure~\ref{fig:teaser} (bottom row) demonstrates the intuition: if we randomly remove some patches from the image then there is a possibility that the dog's face, which is the most discriminative, will not be visible to the model. In this case, the model must seek other relevant parts like the tail and legs in order to do well on the classification task.  By randomly hiding different patches in each training epoch, the model sees different parts of the image and is forced to focus on multiple relevant parts of the object beyond just the most discriminative one.   Importantly, we only apply this random hiding of patches during training and not during testing.  Since the full image is observed during testing, the data distribution will be different to that seen during training.  We show that setting the hidden pixels' value to be the data mean can allow the two distributions to match, and provide a theoretical justification.

Since Hide-and-Seek only alters the input image, it can easily be generalized to different neural networks and tasks.  In this work, we demonstrate its applicability on AlexNet~\cite{krizhevsky-nips2012} and GoogLeNet~\cite{Szegedy-CVPR2015}, and apply the idea to weakly-supervised object localization in images and weakly-supervised action localization in videos.  For the temporal action localization task (in which the start and end times of an action need to be found), random frame sequences are hidden while training a network on action classification, which forces the network to learn the relevant frames corresponding to an action.

\vspace{-10pt}
\paragraph{Contributions.} Our work has three main contributions: 1) We introduce the idea of Hide-and-Seek for weakly-supervised localization and produce state-of-the-art object localization results on the ILSVRC dataset~\cite{Russakovsky-IJCV2015}; 2) We demonstrate the generalizability of the approach on different networks and layers; 3) We extend the idea to the relatively unexplored task of weakly-supervised temporal action localization. 
\section{Related Work}

\paragraph{Weakly-supervised object localization.}
Fully-supervised convolutional networks (CNNs) have demonstrated great performance on object detection~\cite{girshick-cvpr2014,girshick-iccv2015,liu-eccv2016}, segmentation~\cite{long-cvpr2015} and attribute localization~\cite{duan-cvpr2012,zhang-cvpr2014,kiapour-eccv2014}, but require expensive human annotations for training (e.g. bounding box for object localization). To alleviate expensive annotation costs, weakly-supervised approaches learn using cheaper labels, for example, image-level labels for predicting an object's location~\cite{weber-eccv2000,fergus-cvpr2003,Crandall-ECCV2006,siva-eccv2012,bilen-bmcv2014,song-nips2014,wang-eccv2014,cinbis-arxiv2015,Oquab-cvpr15,zhou-cvpr2016}. 

Most weakly-supervised object localization approaches mine discriminative features or patches in the data that frequently appear in one class and rarely in other classes~\cite{weber-eccv2000,fergus-cvpr2003,Crandall-ECCV2006,siva-eccv2012,bilen-bmcv2014,cinbis-cvpr2014,song-icml2014,song-nips2014,cinbis-arxiv2015}. However, these approaches tend to focus only on the most discriminative parts, and thus fail to cover the entire spatial extent of an object.  In our approach, we hide image patches (randomly) during training, which forces our model to focus on multiple parts of an object and not just the most discriminative ones.   Other methods use additional motion cues from weakly-labeled videos to improve object localization~\cite{prest-cvpr2012,krishna-cvpr2016}.  While promising, such videos are not always readily available and can be challenging to obtain especially for static objects.  In contrast, our method does not require any additional data or annotations.

Recent work modify CNN architectures designed for image classification so that the convolutional layers learn to localize objects while performing image classification~\cite{Oquab-cvpr15,zhou-cvpr2016}.  Other network architectures have been designed for weakly-supervised object detection~\cite{Jaderberg-nips2015,Bilen-cvpr2016,kantorov-eccv2016}.  Although these methods have significantly improved the state-of-the-art, they still essentially rely on a classification objective and thus can fail to capture the full extent of an object if the less discriminative parts do not help improve classification performance.  We also rely on a classification objective. However, rather than modifying the CNN architecture, we instead modify the \emph{input image} by hiding random patches from it.  We demonstrate that this enforces the network to give attention to the less discriminative parts and ultimately localize a larger extent of the object.

\vspace{-10pt}
\paragraph{Masking pixels or activations.}  Masking image patches has been applied for object localization~\cite{bazzani_wacv2016}, self-supervised feature learning~\cite{pathak-cvpr2016}, semantic segmentation~\cite{bharath-eccv2014,dai-cvpr2015}, generating
hard occlusion training examples for object detection~\cite{wang-cvpr2017}, and to visualize and understand what a CNN has learned~\cite{Zeiler-eccv2014}.  In particular, for object localization,~\cite{Zeiler-eccv2014,bazzani_wacv2016} train a CNN for image classification and then localize the regions whose masking leads to a large drop in classification performance. Since these approaches mask out the image regions only during \emph{testing} and not during training, the localized regions are limited to the highly-discriminative object parts. In our approach, image regions are masked during \emph{training}, which enables the model to learn to focus on even the less discriminative object parts. Finally, our work is closely related to the adversarial erasing method of~\cite{wei-cvpr2017}, which iteratively trains a sequence of models for weakly-supervised semantic segmentation.  Each model identifies the relevant object parts conditioned on the previous iteration model's output.  In contrast, we only train a single model once---and is thus less expensive---and do not rely on saliency detection to refine the localizations as done in~\cite{wei-cvpr2017}.

\begin{figure*}[t!]
\centering
    \includegraphics[width=0.99\textwidth]{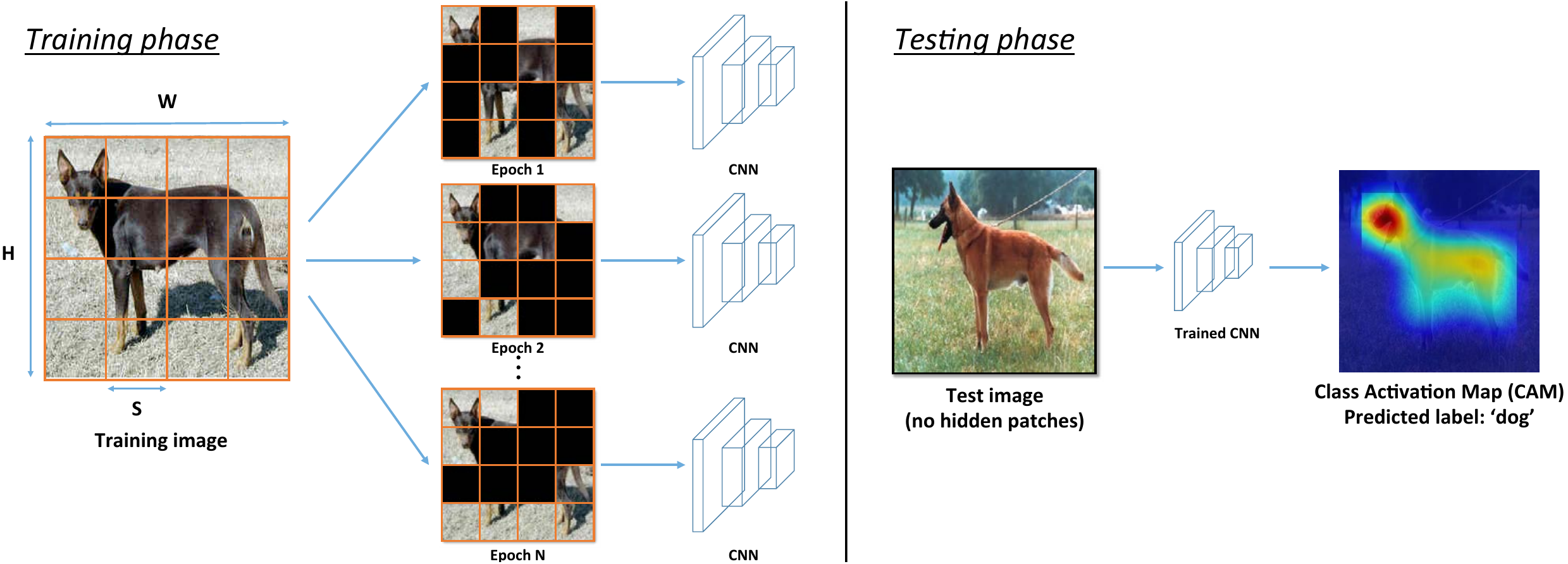}
    \caption{
    \textbf{Approach overview.}
    \textit{Left:} For each training image, we divide it into a grid of $S \times S$ patches.  Each patch is then randomly hidden with probability $p_{hide}$ and given as input to a CNN to learn image classification.  The hidden patches change randomly across different epochs. \textit{Right:} During testing, the full image without any hidden patches is given as input to the trained network.}
\label{fig:pipeline}
\vspace*{-0.1in}
\end{figure*}

Dropout~\cite{srivastava-jmlr2014} and its variants~\cite{wan-icml2013,tompson-cvpr2015} are also related.  There are two main differences: (1) these methods are designed to prevent overfitting while our work is designed to improve localization; and (2) in dropout, units in a layer are dropped randomly, while in our work, contiguous image regions or video frames are dropped.  We demonstrate in the experiments that our approach produces significantly better localizations compared to dropout.

\vspace{-10pt}
\paragraph{Action localization.}
Action localization is a well studied problem~\cite{laptev-cvpr2008,chen-cvpr2012,wang-iccv2013,jhuang-iccv2013,kantorov-cvpr2014}.  Recent CNN-based approaches~\cite{yeung-cvpr2016,Shou-CVPR2016} have shown superior performance compared to previous hand-crafted approaches. These fully-supervised methods require the start and end time of an action in the video during the training to be annotated, which can be expensive to obtain.  Weakly-supervised approaches learn from movie scripts~\cite{laptev-cvpr2008,duchenne-iccv2009} or an ordered list of actions~\cite{bojanowski-eccv2014,huang-eccv2016}.  Sun et al.~\cite{sun-mm2016} combine weakly-labeled videos with web images for action localization. In contrast to these approaches, our approach only uses a single video-level action label for temporal action localization. \cite{gan-cvpr2015} also only use video-level action labels for action localization with the focus on finding the key event frames of an action. We instead focus on localizing the full extent of an action.

\section{Approach}

In this section, we first describe our Hide-and-Seek algorithm for object localization in images followed by action localization in videos.

\subsection{Weakly-supervised object localization}

For weakly-supervised object localization, we are given a set of images $I_{set} = \{I_1, I_2,.....,I_N\}$ in which each image $I$ is labeled only with its category label.  Our goal is to learn an object localizer that can predict both the category label as well as the bounding box for the object-of-interest in a new test image $I_{test}$. In order to learn the object localizer, we train a CNN which simultaneously learns to localize the object while performing the image classification task.  While numerous approaches have been proposed to solve this problem, existing methods (e.g.,~\cite{song-icml2014,cinbis-arxiv2015,Oquab-cvpr15,zhou-cvpr2016}) are prone to localizing only the most discriminative object parts, since those parts are sufficient for optimizing the classification task.

To enforce the network to learn all of the relevant parts of an object, our key idea is to randomly hide patches of each input image $I$ during training, as we explain next.  

\vspace{-10pt}
\paragraph{Hiding random image patches.}

The purpose of hiding patches is to show different parts of an object to the network while training it for the classification task.  By hiding patches randomly, we can ensure that the most discriminative parts of an object are not always visible to the network, and thus \emph{force} it to also focus on other relevant parts of the object.  In this way, we can overcome the limitation of existing weakly-supervised methods that focus only on the most discriminative parts of an object.

Concretely, given a training image $I$ of size $W \times H \times 3$, we first divide it into a grid with a fixed patch size of $S \times S \times 3$.  This results in a total of $(W \times H)/(S \times S)$ patches.  We then hide each patch with $p_{hide}$ probability.  For example, in Fig.~\ref{fig:pipeline} left, the image is of size $224 \times 224 \times 3$, and it is divided into $16$ patches of size $56 \times 56 \times 3$.  Each patch is hidden with $p_{hide}=0.5$ probability.  We take the new image $I'$ with the hidden patches, and feed it as a training input to a CNN for classification.

Importantly, for each image, we randomly hide a different set of patches.  Also, for the same image, we randomly hide a different set of patches in each training epoch.  This property allows the network to learn multiple relevant object parts for each image.  For example, in Fig.~\ref{fig:pipeline} left, the network sees a different $I'$ in each epoch due to the randomness in hiding of the patches. In the first epoch, the dog's face is hidden while its legs and tail are clearly visible.  In contrast, in the second epoch, the face is visible while the legs and tail are hidden.  Thus, the network is forced to learn all of the relevant parts of the dog rather than only the highly discriminative part (i.e., the face) in order to perform well in classifying the image as a `dog'.  

We hide patches only during training.  During testing, the full image---without any patches hidden---is given as input to the network; Fig.~\ref{fig:pipeline} right.  Since the network has learned to focus on multiple relevant parts during training, it is not necessary to hide any patches during testing.  This is in direct contrast to~\cite{bazzani_wacv2016}, which hides patches during testing but not during training.  For~\cite{bazzani_wacv2016}, since the network has already learned to focus on the most discimirinative parts during training, it is essentially too late, and hiding patches during testing has no significant effect on localization performance.

\begin{figure}[t!]
\centering
    \includegraphics[width=0.45\textwidth]{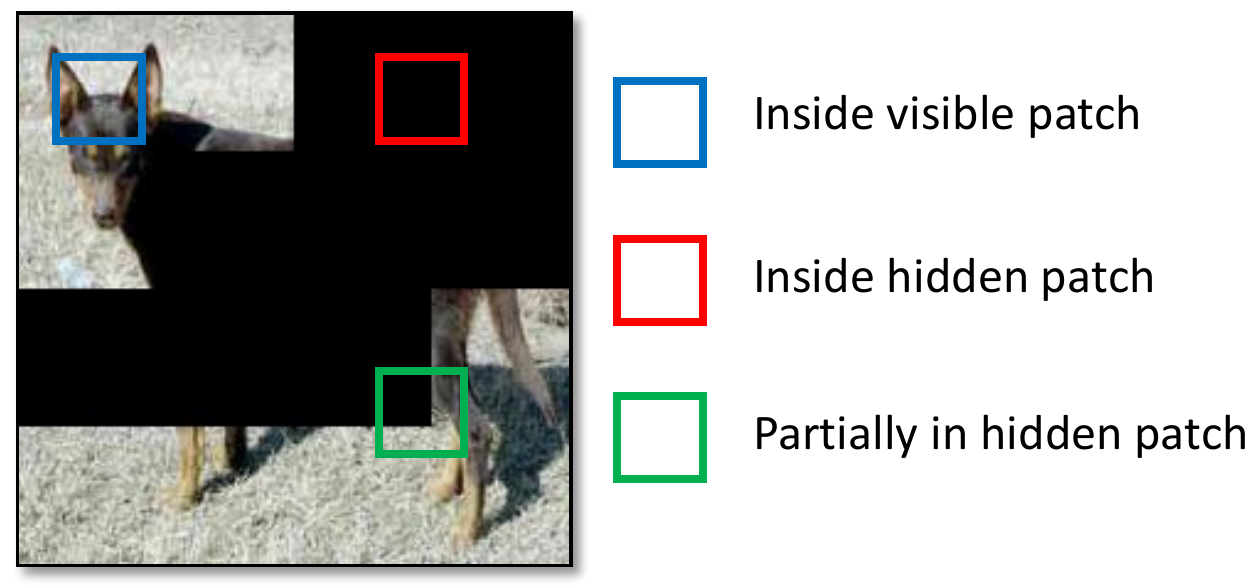}
    \caption{There are three types of convolutional filter activations after hiding patches: a convolution filter can be completely within a visible region (blue box), completely within a hidden region (red box), or partially within a visible/hidden region (green box).}
\label{fig:scaling}
\vspace*{-0.1in}
\end{figure}

\vspace{-10pt}
\paragraph{Setting the hidden pixel values.}\label{sec:scaling}

There is an important detail that we must be careful about.  Due to the discrepancy of hiding patches during training while not hiding patches during testing, the first convolutional layer activations during training versus testing will have different distributions.  For a trained network to generalize well to new test data, the activation distributions should be roughly equal.  That is, for any unit in a neural network that is connected to $\mathbf{x}$ units with $\mathbf{w}$ outgoing weights, the distribution of $\mathbf{w}^\top\mathbf{x}$ should be roughly the same during training and testing.  However, in our setting, this will not necessarily be the case since some patches in each training image will be hidden while none of the patches in each test image will ever be hidden.

Specifically, in our setting, suppose that we have a convolution filter $F$ with kernel size $K \times K$ and three-dimensional weights $W = \{\mathbf{w}_1,\mathbf{w}_2,....,\mathbf{w}_{k\times k}\}$, which is applied to an RGB patch $X= \{\mathbf{x}_1,\mathbf{x}_2,....,\mathbf{x}_{k \times k}\}$ in image $I'$.  Denote $\mathbf{v}$ as the vector representing the RGB value of every hidden pixel.  There are three types of activations:
\begin{enumerate}
  \item $F$ is completely within a visible patch (Fig.~\ref{fig:scaling}, blue box).  The corresponding output will be $\sum_{i=1}^{k \times k} \mathbf{w}_i^\top \mathbf{x}_i$.
  \item $F$ is completely within a hidden patch (Fig.~\ref{fig:scaling}, red box).  The corresponding output will be $\sum_{i=1}^{k \times k} \mathbf{w}_i^\top \mathbf{v}$.
  \item $F$ is partially within a hidden patch (Fig.~\ref{fig:scaling}, green box).  The corresponding output will be $\sum_{m \in visible} \mathbf{w}_m^\top \mathbf{x}_m + \sum_{n \in hidden} \mathbf{w}_n^\top \mathbf{v}$.
\end{enumerate}

During testing, $F$ will always be completely within a visible patch, and thus its output will be $\sum_{i=1}^{k \times k} \mathbf{w}_i^\top \mathbf{x}_i$.  This matches the expected output during training in only the first case.  For the remaining two cases, when $F$ is completely or partially within a hidden patch, the activations will have a distribution that is different to those seen during testing.

We resolve this issue by setting the RGB value $\mathbf{v}$ of a hidden pixel to be equal to the mean RGB vector of the images over the entire dataset: $\mathbf{v} = \mu = \frac{1}{N_{pixels}} \sum_j \mathbf{x}_j$, where $j$ indexes all pixels in the entire training dataset and $N_{pixels}$ is the total number of pixels in the dataset.  Why would this work?  Essentially, we are assuming that in expectation, the output of a patch will be equal to that of an average-valued patch: $\mathbb{E} [\sum_{i=1}^{k \times k} \mathbf{w}_i^\top \mathbf{x}_i] = \sum_{i=1}^{k \times k} \mathbf{w}_i^\top \mu$.  By replacing $\mathbf{v}$ with $\mu$, the outputs of both the second and third cases will be $\sum_{i=1}^{k \times k} \mathbf{w}_i^\top \mu$, and thus will match the expected output during testing (i.e., of a fully-visible patch).\footnote{For the third case: $\sum_{m \in visible} \mathbf{w}_m^\top \mathbf{x}_m + \sum_{n \in hidden} \mathbf{w}_n^\top \mu \approx \sum_{m \in visible} \mathbf{w}_m^\top \mu + \sum_{n \in hidden} \mathbf{w}_n^\top \mu = \sum_{i=1}^{k \times k} \mathbf{w}_i^\top \mu$.}

This process is related to the scaling procedure in dropout~\cite{srivastava-jmlr2014}, in which the outputs are scaled proportional to the drop rate during testing to match the expected output during training.  In dropout, the outputs are dropped uniformly across the entire feature map, independently of spatial location.  If we view our hiding of the patches as equivalent to ``dropping'' units, then in our case, we cannot have a global scale factor since the output of a patch depends on whether there are any hidden pixels.  Thus, we instead set the hidden values to be the expected pixel value of the training data as described above, and do not scale the corresponding output.  Empirically, we find that setting the hidden pixel in this way is crucial for the network to behave similarly during training and testing.

\vspace{-10pt}
\paragraph{Object localization network architecture.}

Our approach of hiding patches is independent of the network architecture and can be used with any CNN designed for object localization.  For our experiments, we choose to use the network of Zhou et al.~\cite{zhou-cvpr2016}, which performs global average pooling (GAP) over the convolution feature maps to generate a class activation map (CAM) for the input image that represents the discriminative regions for a given class.  This approach has shown state-of-the-art performance for the ILSVRC localization challenge~\cite{Russakovsky-IJCV2015} in the weakly-supervised setting, and existing CNN architectures like AlexNet~\cite{krizhevsky-nips2012} and GoogLeNet~\cite{Szegedy-CVPR2015} can easily be modified to generate a CAM.

To generate a CAM for an image, global average pooling is performed after the last convolutional layer and the result is given to a classification layer to predict the image's class probabilities.  The weights associated with a class in the classification layer represent the importance of the last convolutional layer's feature maps for that class.  More formally, denote $F=\{F_1,F_2,..,F_M\}$ to be the $M$ feature maps of the last convolutional layer and $W$ as the $N \times M$ weight matrix of the classification layer, where $N$ is number of classes.  Then, the CAM for class $c$ for image $I$ is:
\vspace{-2pt}
\begin{equation}
CAM(c,I) = \sum_{i=1}^{M} W(c,i) \cdot F_i(I).
\vspace{-2pt}
\label{eq:cam}
\end{equation}

Given the CAM for an image, we generate a bounding box using the method proposed in~\cite{zhou-cvpr2016}. Briefly, we first threshold the CAM to produce a binary foreground/background map, and then find connected components among the foreground pixels. Finally, we fit a tight bounding box to the largest connected component.  We refer the reader to~\cite{zhou-cvpr2016} for more details.

\subsection{Weakly-supervised action localization}

Given a set of untrimmed videos $V_{set}=\{V_1,V_2,...,V_N\}$ and video class labels, our goal here is to learn an action localizer that can predict the label of an action as well as its start and end time for a test video $V_{test}$.  Again the key issue is that for any video, a network will focus mostly on the highly-discriminative frames in order to optimize classification accuracy instead of identifying all relevant frames.  Inspired by our idea of hiding the patches in images, we propose to hide frames in videos to improve action localization.

Specifically, during training, we uniformly sample video $F_{total}$ frames from each videos. We then divide the $F_{total}$ frames into continuous segments of fixed size $F_{segment}$; i.e., we have $F_{total} / F_{segemnt}$ segments. Just like with image patches, we hide each segment with probability $p_{hide}$ before feeding it into a deep action localizer network.  We generate class activation maps (CAM) using the procedure described in the previous section.  In this case, our CAM is a one-dimensional map representing the discriminative frames for the action class. We apply thresholding on this map to obtain the start and end times for the action class.

\section{Experiments}\label{section:results}

We perform quantitative and qualitative evaluations of Hide-and-Seek for object localization in images and action localization in videos.  We also perform ablative studies to compare the different design choices of our algorithm.

\vspace{-10pt}
\paragraph{Datasets and evaluation metrics.} We use ILSVRC 2016~\cite{Russakovsky-IJCV2015} to evaluate object localization accuracy. For training, we use 1.2 million images with their class labels (1000 categories). We compare our approach with the baselines on the validation data. We use three evaluation metrics to measure performance: 1) Top-1 localization accuracy (\emph{Top-1 Loc}): fraction of images for which the predicted class with the highest probability is the same as the ground-truth class \emph{and} the predicted bounding box for that class has more than $50\%$ IoU with the ground-truth box. 2) Localization accuracy with known ground-truth class (\emph{GT-known Loc}): fraction of images for which the predicted bounding box for the ground-truth class has more than $50\%$ IoU with the ground-truth box.  As our approach is primarily designed to improve localization accuracy, we use this criterion to measure localization accuracy independent of classification performance. 3) We also use classification accuracy (\emph{Top-1 Clas}) to measure the impact of Hide-and-Seek on image classification performance.

For action localization, we use THUMOS 2014 validation data~\cite{jiang-14}, which consists of 1010 untrimmed videos belonging to 101 action classes. We train over all untrimmed videos for the classification task and then evaluate localization on the 20 classes that have temporal annotations. Each video can contain multiple instances of a class. For evaluation we compute mean average precision (mAP), and consider a prediction to be correct if it has IoU $> \theta$ with ground-truth. We vary $\theta$ to be 0.1, 0.2, 0.3, 0.4, and 0.5. As we are focusing on localization ability of the network, we assume we know the ground-truth class label of the video.

\vspace{-10pt}
\paragraph{Implementation details.} To learn the object localizer, we use the same modified AlexNet and GoogLeNet networks introduced in~\cite{zhou-cvpr2016} (AlexNet-GAP and  GoogLeNet-GAP).  AlexNet-GAP is identical to AlexNet until pool5 (with stride 1) after which two new conv layers are added.  Similarly for GoogLeNet-GAP, layers after inception-4e are removed and a single conv layer is added.  For both AlexNet-GAP and GoogLeNet-GAP, the output of the last conv layer goes to a global average pooling (GAP) layer, followed by a softmax layer for classification.  Each added conv layer has 512 and 1024 kernels of size $3 \times 3$, stride 1, and pad 1 for AlexNet-GAP and  GoogLeNet-GAP, respectively. 

We train the networks from scratch for 55 and 40 epochs for AlexNet-GAP and GoogLeNet-GAP, respectively, with a batch size of 128 and initial learning rate of 0.01. We gradually decrease the learning rate to 0.0001.  We add batch normalization~\cite{bn} after every conv layer to help convergence of GoogLeNet-GAP. For simplicity, unlike the original AlexNet architecture~\cite{krizhevsky-nips2012}, we do not group the conv filters together (it produces statistically the same \emph{Top-1 Loc} accuracy as the grouped version for both AlexNet-GAP but has better image classification performance). The network remains exactly the same with (during training) and without (during testing) hidden image patches. To obtain the binary fg/bg map, $20\%$ and $30\%$ of the max value of the CAM is chosen as the threshold for AlexNet-GAP and GoogLeNet-GAP, respectively; the thresholds were chosen by observing a few qualitative results on training data.  During testing, we average 10 crops (4 corners plus center, and same with horizontal flip) to obtain class probabilities and localization maps.  We find similar localization/classification performance when fine-tuning pre-trained networks.

For action localization, we compute C3D~\cite{tran-iccv2015} fc7 features using a model pre-trained on Sports 1 million~\cite{karpathy-CVPR14}. We compute 10 feats/sec (each feature is computed over 16 frames) and uniformly sample 2000 features from the video. We then divide the video into 20 equal-length segments each consisting of $F_{segment} = 100$ features.  During training, we hide each segment with $p_{hide} = 0.5$.  For action classification, we feed C3D features as input to a CNN with two conv layers followed by a global max pooling and softmax classification layer. Each conv layer has 500 kernels of size $1 \times 1$, stride 1. For any hidden frame, we assign it the dataset mean C3D feature. For thresholding, $50\%$ of the max value of the CAM is chosen. All continuous segments after thresholding are considered predictions.

\begin{figure*}[t!]
\centering
    \includegraphics[width=0.94\textwidth]{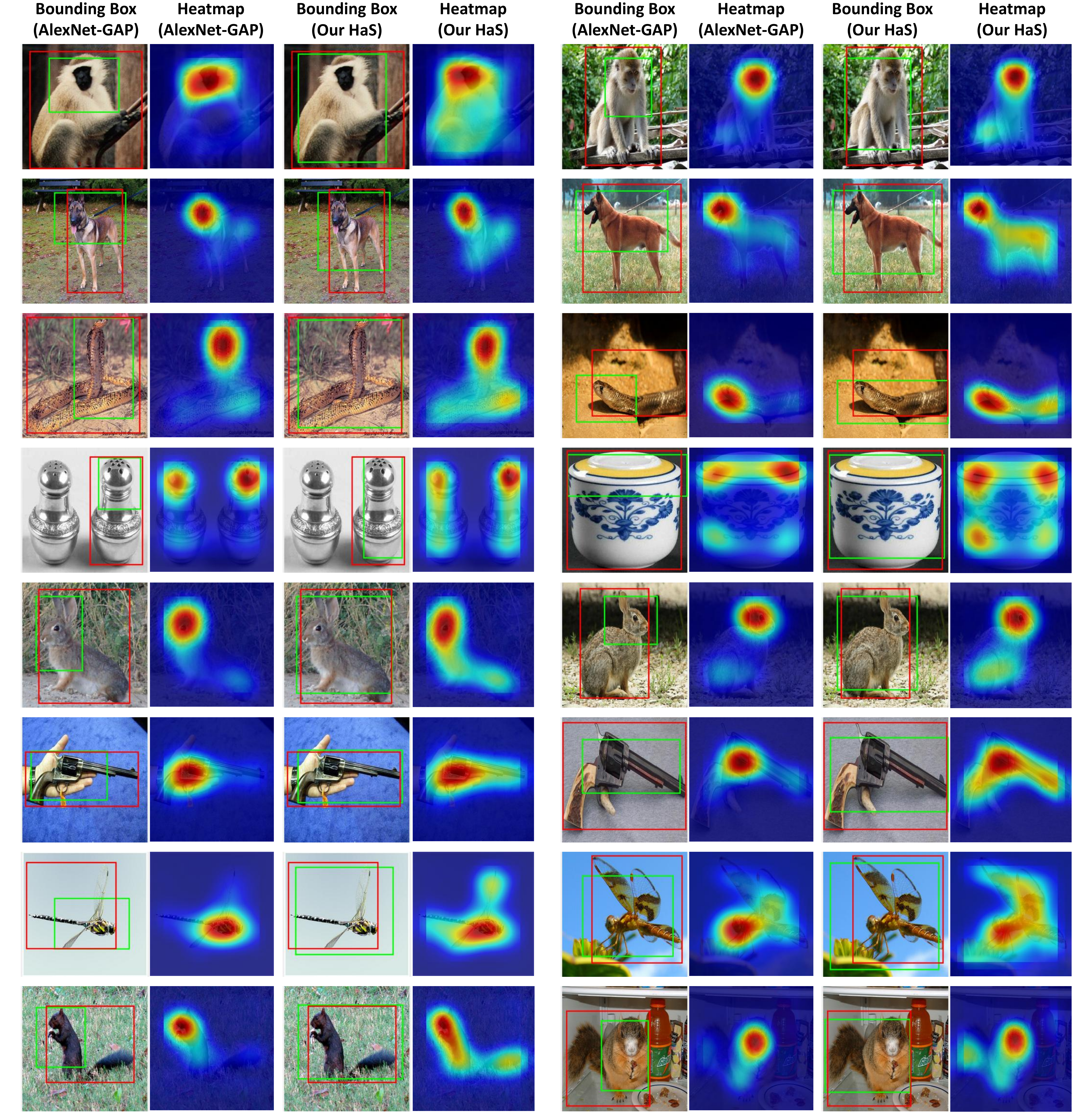}
    \caption{Qualitative object localization results.  We compare our approach with AlexNet-GAP~\cite{zhou-cvpr2016} on the ILVRC validation data. For each image, we show the bounding box and CAM obtained by AlexNet-GAP (left) and our method (right).  Our Hide-and-Seek approach localizes multiple relevant parts of an object whereas AlexNet-GAP mainly focuses only on the most discriminative parts.}
    \vspace*{-0.12in}
\label{fig:qualresults}
\end{figure*}

\footnotetext[2]{\cite{zhou-cvpr2016} does not provide GT-known loc, so we compute on our own GAP implementations, which achieve similar Top-1 Loc accuracy.}

\begin{table}[t!]
\begin{center}
    \footnotesize
    \begin{tabular}{| c | c | c| c|}
    	
    \hline    	
    Methods & GT-known Loc &  Top-1 Loc  & Top-1 Clas   \\
    \hline
    
    AlexNet-GAP~\cite{zhou-cvpr2016}  &  54.90\footnotemark[2] & 36.25 & \textbf{60.23}\\
    AlexNet-HaS-16     & 57.86 & 36.77 & 57.97  \\
    AlexNet-HaS-32            & \textbf{58.75} & 37.33 & 57.94 \\
    AlexNet-HaS-44            & 58.55 & 37.54 & 58.10 \\
    AlexNet-HaS-56            & 58.43 & 37.34 & 58.13  \\
    AlexNet-HaS-Mixed       & 58.68 & \textbf{37.65} & 58.68 \\

            \hline
     GoogLeNet-GAP~\cite{zhou-cvpr2016}  & 58.41\footnotemark[2] & 43.60 & \textbf{71.95}  \\
     GoogLeNet-HaS-16            & 59.83 & 44.62 & 70.49 \\
                GoogLeNet-HaS-32            &  \textbf{60.29} & \textbf{45.21} & 70.70 \\
                GoogLeNet-HaS-44          & 60.11 & 44.75 & 70.34  \\
                GoogLeNet-HaS-56            & 59.93 & 44.78 & 70.37  \\
                \hline

    \end{tabular}
    \caption{Localization accuracy on ILSVRC validation data with different patch sizes for hiding.  Our Hide-and-Seek always performs better than AlexNet-GAP~\cite{zhou-cvpr2016}, which sees the full image.}
    \label{table:patch_size_results}
\end{center}
\vspace*{-0.15in}
\end{table}

\subsection{Object localization quantitative results}

We first analyze object localization accuracy on the ILSVRC validation data. Table~\ref{table:patch_size_results} shows the results using the \emph{Top-1 Loc} and \emph{GT-known Loc} evaluation metrics.  AlexNet-GAP~\cite{zhou-cvpr2016} is our baseline in which the network has seen the full image during training without any hidden patches. Alex-HaS-N is our approach, in which patches of size $N \times N$ are hidden with 0.5 probability during training.

\vspace{-10pt}
\paragraph{Which patch size $N$ should we choose?} We explored four different patch sizes $N = \{16, 32, 44, 56\}$, and each performs significantly better than AlexNet-GAP for both \emph{GT-known Loc} and \emph{Top-1 Loc}. Our GoogLeNet-HaS-N models also outperfors GoogLeNet-GAP for all patch sizes.  These results clearly show that hiding patches during training leads to better localization. Although our approach can lose some classification accuracy (\emph{Top-1 Clas}) since it has never seen a complete image and thus may not have learned to relate certain parts, the huge boost in localization performance (which can be seen by comparing the \emph{GT-known Loc} accuracies) makes up for any potential loss in classification.

\begin{table}[t!]
\begin{center}
    \footnotesize
    \begin{tabular}{| c | c | c| c|}
    \hline    	
    Methods & GT-known Loc &  Top-1 Loc    \\
    \hline
    Backprop on AlexNet~\cite{simonyan-iclr2014} & - & 34.83 \\
    AlexNet-GAP~\cite{zhou-cvpr2016}  &  54.90 & 36.25\\
    
    Ours         & \textbf{58.68} & \textbf{37.65}  \\
    \hline
    AlexNet-GAP-ensemble   & 56.91 & 38.58 \\
    Ours-ensemble       & \textbf{60.14} & \textbf{40.40}   \\
            \hline
     Backprop on GoogLeNet~\cite{simonyan-iclr2014} & - & 38.69 \\
         GoogLeNet-GAP~\cite{zhou-cvpr2016} & 58.41 & 43.60 \\
         Ours         &  \textbf{60.29} & \textbf{45.21} \\
     \hline
    \end{tabular}
    \caption{Localization accuracy on ILSVRC val data compared to state-of-the-art.  Our method outperforms all previous methods.}
    \label{table:main_results}
\end{center}
\vspace*{-0.15in}
\end{table}

We also train a network (AlexNet-HaS-Mixed) with mixed patch sizes. During training, for each image in every epoch, the patch size $N$ to hide is chosen randomly from 16, 32, 44 and 56 as well as no hiding (full image).  Since different sized patches are hidden, the network can learn complementary information about different parts of an object (e.g. small/large patches are more suitable to hide smaller/larger parts). Indeed, we achieve the best results for \emph{Top-1 Loc} using AlexNet-HaS-Mixed. 

\vspace{-10pt}
\paragraph{Comparison to state-of-the-art.}  Next, we choose our best model for AlexNet and GoogLeNet, and compare it with state-of-the-art methods on ILSVRC validation data; see Table~\ref{table:main_results}. Our method performs 3.78\% and 1.40\% points better than AlexNet-GAP~\cite{zhou-cvpr2016} on \emph{GT-known Loc} and \emph{Top-1 Loc}, respectively. For GoogLeNet, our model gets a boost of 1.88\% and 1.61\% points compared to GoogLeNet-GAP for \emph{GT-known Loc} and \emph{Top-1 Loc} accuracy, respectively.  Importantly, these gains are obtained simply by changing the input image without changing the network architecture.

\vspace{-10pt}
\paragraph{Ensemble model.}  Since each patch size provides complementary information (as seen in the previous section), we also create an ensemble model of different patch sizes (Ours-ensemble).  To produce the final localization for an image, we average the CAMs obtained using AlexNet-HaS-16, 32, 44, and 56, while for classification, we average the classification probabilities of all four models as well as the probability obtained using AlexNet-GAP.  This ensemble model gives a boost of 5.24 \% and 4.15\% over AlexNet-GAP for \emph{GT-known Loc} and \emph{Top-1 Loc}, respectively.  For a more fair comparison, we also combine the results of five independent AlexNet-GAPs to create an ensemble baseline.  Ours-ensemble outperforms this strong baseline (AlexNet-GAP-ensemble) by 3.23\% and 1.82\% for \emph{GT-known Loc} and \emph{Top-1 Loc}, respectively.

\subsection{Object localization qualitative results}

In Fig.~\ref{fig:qualresults}, we visualize the class activation map (CAM) and bounding box obtained by our AlexNet-HaS approach versus those obtained with AlexNet-GAP.  In each image pair, the first image shows the predicted (green) and ground-truth (red) bounding box. The second image shows the CAM, i.e., where the network is focusing for that class.  Our approach localizes more relevant parts of an object compared to AlexNet-GAP and is not confined to only the most discriminative parts. For example, in the first, second, and fifth rows AlexNet-GAP only focuses on the face of the animals, whereas our method also localizes parts of the body.  Similarly, in the third and last rows AlexNet-GAP misses the tail for the snake and squirrel while ours gets the tail. 

\subsection{Further Analysis of Hide-and-Seek}

\paragraph{Comparison with dropout.}  Dropout~\cite{srivastava-jmlr2014} has been extensively used to reduce overfitting in deep network.  Although it is not designed to improve localization, the dropping of units is related to our hiding of patches. We therefore conduct an experiment in which 50\% dropout is applied at the image layer. We noticed that the due to the large dropout rate at the pixel-level, the learned filters develop a bias toward a dropped-out version of the images and produces significantly inferior classification and localization performance (AlexNet-dropout-trainonly). If we also do dropout during testing (AlexNet-dropout-traintest) then performance improves but is still much lower compared to our approach (Table~\ref{table:dropout_results}).   Since dropout drops pixels (and RGB channels) randomly, information from the most relevant parts of an object will still be seen by the network with high probability, which makes it likely to focus on only the most discriminative parts.

\begin{table}[t!]
            \begin{center}
                \footnotesize
                \begin{tabular}{| c | c | c|}
                \hline    	
                Methods & GT-known Loc &  Top-1 Loc \\
                \hline

                Ours      &     \textbf{58.68} & \textbf{37.65}  \\
                AlexNet-dropout-trainonly            & 42.17 & 7.65  \\
                AlexNet-dropout-traintest           &  53.48  & 31.68  \\

                \hline
                 \end{tabular}
                        \caption{Our approach outperforms Dropout~\cite{srivastava-jmlr2014} for localization.}
                        \label{table:dropout_results}
                        \end{center}
                        \vspace*{-0.15in}
                        \end{table}

\begin{table}[t!]
        \begin{center}
            \footnotesize
            \begin{tabular}{| c | c | c|}
            \hline    	
            Methods & GT-known Loc &  Top-1 Loc \\
            \hline
            AlexNet-GAP            & 54.90 & 36.25 \\
            AlexNet-Avg-HaS            & 58.43 & 37.34   \\
            AlexNet-GMP            & 50.40 & 32.52  \\
            AlexNet-Max-HaS            & \textbf{59.27} &  \textbf{37.57}  \\

            \hline
             \end{tabular}
                    \caption{Global average pooling (GAP) vs.~global max pooling (GMP).  Unlike~\cite{zhou-cvpr2016}, for Hide-and-Seek GMP still performs well for localization. For this experiment, we use patch size 56.}
                    \label{table:max_results}
                    \end{center}
                    \vspace*{-0.15in}
                    \end{table}

\vspace{-10pt}
\paragraph{Do we need global average pooling?}  \cite{zhou-cvpr2016} showed that GAP is better than global max pooling (GMP) for object localization, since average pooling encourages the network to focus on all the discriminative parts.  For max pooling, only the most discriminative parts need to contribute. But is global max pooling hopeless for localization?

With our Hide-and-Seek, even with max pooling, the network is forced to focus on a different discriminative parts.   In Table~\ref{table:max_results}, we see that max pooling (AlexNet-GMP) is inferior to average poling (AlexNet-GAP) for the baselines. But with Hide-and-Seek, max pooling (AlexNet-Max-HaS) localization accuracy increases by a big margin and even slightly outperforms average pooling (AlexNet-Avg-HaS). The slight improvement is likely due to max pooling being more robust to noise.

\begin{table}[t!]
                \begin{center}
                    \footnotesize
                    \begin{tabular}{| c | c | c|}
                    \hline    	
                    Methods & GT-known Loc &  Top-1 Loc \\
                    \hline

                    AlexNet-GAP            & 54.90 & 36.25 \\
                    AlexNet-HaS-conv1-5            & 57.36 & 36.91  \\
                    AlexNet-HaS-conv1-11            &  \textbf{58.33}  & \textbf{37.38}  \\

                    \hline
                     \end{tabular}
                            \caption{Applying Hide-and-Seek to the first conv layer. The improvement over~\cite{zhou-cvpr2016} shows the generality of the idea.}
                            \label{table:conv_results}
                            \end{center}
                            \vspace*{-0.15in}
                            \end{table}

\vspace{-10pt}
\paragraph{Hide-and-Seek in convolutional layers.} We next apply our idea to convolutional layers.  We divide the convolutional feature maps into a grid and hide each patch (and all of its corresponding channels) with 0.5 probability.  We hide patches of size 5 (AlexNet-HaS-conv1-5) and 11 (AlexNet-HaS-conv1-11) in the conv1 feature map (which has size $ 55 \times 55 \times 96$).  Table~\ref{table:conv_results} shows that this leads to a big boost in performance compared to the baseline AlexNet-GAP. This shows that our idea of randomly hiding patches can be generalized to the convolutional layers.

\begin{table}[t!]
              \begin{center}
                  \footnotesize
                  \begin{tabular}{| c | c | c|}
                  \hline    	
                  Methods & GT-known Loc &  Top-1 Loc\\
                  \hline

                  AlexNet-HaS-25\%            & 57.49 & 37.77  \\
                  AlexNet-HaS-33\%          & 58.12 & 38.05  \\
                  AlexNet-HaS-50\%            & 58.43 & 37.34  \\
                  AlexNet-HaS-66\%            &  58.52  & 35.72  \\
                  AlexNet-HaS-75\%             &  58.28  & 34.21  \\

                  \hline
                   \end{tabular}
                          \caption{Varying the hiding probability. Higher probabilities lead to decrease in \emph{Top-1 Loc} whereas lower probability leads to smaller \emph{GT-known Loc}. For this experiment, we use patch size 56.}
                          \label{table:drop_percent_results}
                          \end{center}
                          \vspace*{-0.13in}
                          \end{table}

\vspace{-10pt}
\paragraph{Probability of hiding.} In all of the previous experiments, we hid patches with 50\% probability. In Table~\ref{table:drop_percent_results}, we measure the \emph{GT-known Loc} and \emph{Top-1 Loc} when we use different hiding probabilities.  If we increase the probability then \emph{GT-known Loc} remains almost the same while \emph{Top-1 Loc} decreases a lot. This happens because the network sees fewer pixels when the hiding probability is high; as a result, classification accuracy reduces and \emph{Top-1 Loc} drops.  If we decrease the probability then \emph{GT-known Loc} decreases but our \emph{Top-1 Loc} improves.  In this case, the network sees more pixels so its classification improves but since less parts are hidden, it will focus more on only the discriminative parts decreasing its localization ability.

\begin{table}[t!]
              \begin{center}
                  \footnotesize
                  \begin{tabular}{| c | c | c| c| c| c|}
                  \hline    	
	                Methods     & IOU thresh = 0.1 & 0.2 & 0.3 & 0.4 & 0.5 \\
	                \hline
					Video-full & 34.23 &   25.68 &   17.72   & 11.00 &   6.11\\
					Video-HaS & \textbf{36.44}   & \textbf{27.84} &   \textbf{19.49} &   \textbf{12.66}  &  \textbf{6.84}\\
					 \hline
					                   \end{tabular}

      \caption{Action localization accuracy on THUMOS validation data.   Across all 5 IoU thresholds, our Video-HaS outperforms the full video baseline (Video-full).}
                          \label{table:frame_hide}
                          \end{center}
                          \vspace*{-0.2in}
                          \end{table}
                          					
\subsection{Action localization results}

Finally, we evaluate action localization accuracy.  We compare our approach (Video-HaS), which randomly hides frame segments while learning action classification, with a baseline that sees the full video (Video-full).  Table~\ref{table:frame_hide} shows the result on THUMOS validation data.  Video-HaS consistently outperforms Video-full for action localization task, which shows that hiding frames forces our network to focus on more relevant frames, which ultimately leads to better action localization.  \emph{We show qualitative results in the supp.}

\section{Conclusion}
\vspace*{-0.05in}

We presented `Hide-and-Seek', a novel weakly-supervised framework to improve object localization in images and temporal action localization in videos.  By randomly hiding patches/frames in a training image/video, we force the network to learn to focus on multiple relevant parts of an object/action.  Our extensive experiments showed improved localization accuracy over state-of-the-art methods.

\vspace{-10pt}
\paragraph{Acknowledgements.}
 This work was supported in part by Intel Corp, Amazon Web Services Cloud Credits for Research, and GPUs donated by NVIDIA.

{\small
\bibliographystyle{ieee}
\bibliography{main}
}

\end{document}